\pdfoutput=1
\documentclass[11pt]{article}

\usepackage[preprint]{acl}

\usepackage{times}
\usepackage{latexsym}
\usepackage{subfigure}
\usepackage[T1]{fontenc}
\usepackage[utf8]{inputenc}

\usepackage{microtype}
\usepackage{booktabs}
\usepackage{enumitem}
\usepackage{subcaption}
\usepackage{multirow}
\usepackage{inconsolata}
\usepackage{graphicx}

\usepackage{xspace} % Ensure this is in the preamble

\newcommand{\eg}{e.g.\@\xspace}

\newcommand{\ie}{i.e.\@\xspace}

\definecolor{lightblue}{rgb}{0.12, 0.44, 0.80}

\title{The Hallucinations Leaderboard -- An Open Effort to \\ Measure Hallucinations in Large Language Models}

\author{Giwon Hong\textsuperscript{1}\thanks{Equal contribution. GH conducted the first draft of the paper and the analyses in Section \ref{sec:comprehensive_alaysis}. APG contributed to the first version of the leaderboard and corresponding \href{https://huggingface.co/blog/leaderboard-hallucinations}{blog post}. RS contributed to the summarisation tasks. XD contributed to the knowledge memorisation tasks. LPB contributed to the writing and experimental design. MR contributed to prompt robustness evaluation and writing. PM and CF created the first version of the leaderboard and corresponding \href{https://huggingface.co/blog/leaderboard-hallucinations}{blog post}.} \ Aryo Pradipta Gema\textsuperscript{1}\footnotemark[1]
\ Rohit Saxena \textsuperscript{1}\footnotemark[1]
\ \textbf{Xiaotang Du}\textsuperscript{1}\footnotemark[1] \ \textbf{Ping Nie}\textsuperscript{5}\footnotemark[1]  \ \textbf{Yu Zhao}\textsuperscript{1}\footnotemark[1] \\ \textbf{Laura Perez-Beltrachini}\textsuperscript{1} \ \textbf{Max Ryabinin}\textsuperscript{4} \ \textbf{Xuanli He}\textsuperscript{3} \ \textbf{Clémentine Fourrier}\textsuperscript{2} \textbf{Pasquale Minervini}\textsuperscript{1}\footnotemark[1]\thanks{Corresponding authors.}\\
  \textsuperscript{1}School of Informatics, University of Edinburgh \quad  \textsuperscript{2} Hugging Face \\
  \textsuperscript{3}Department of Computer Science, University College London \quad \textsuperscript{4}Together AI \\
  \textsuperscript{5}School of Electronics Engineering and Computer Science, Peking University\\
  \texttt{ \{first.last, lperez, p.minervini\}@ed.ac.uk}\qquad \texttt{mryabinin0@gmail.com} \\
  \texttt{clementine@huggingface.co}\qquad \texttt{xuanli.he@ucl.ac.uk} \qquad \texttt{ping.nie@pku.edu.cn}\\
  }

\begin{document}
\maketitle
\begin{abstract}
Large Language Models (LLMs) have transformed the Natural Language Processing (NLP) landscape with their remarkable ability to understand and generate human-like text.
However, these models are prone to ``hallucinations'' --- outputs that do not align with factual reality or the input context.
This paper introduces the Hallucinations Leaderboard, an open initiative to quantitatively measure and compare the tendency of each model to produce hallucinations.
The leaderboard uses a comprehensive set of benchmarks focusing on different aspects of hallucinations, such as factuality and faithfulness, across various tasks, including question-answering, summarisation, and reading comprehension.
Our analysis provides insights into the performance of different models, guiding researchers and practitioners in choosing the most reliable models for their applications.
\end{abstract}

\section{Introduction}
\label{sec:introduction}
Large Language Models (LLMs) have emerged as powerful language generators, \ie generating fluent and topically coherent text, and few-shot task instruction followers \cite{Radford2019LanguageMA,NEURIPS2020_1457c0d6,wei2022finetuned,NEURIPS2022_b1efde53,10.1145/3560815}. Because they are trained on large amounts of textual data, they are also a prominent source of knowledge \cite{petroni-etal-2019-language,roberts-etal-2020-much,safavi-koutra-2021-relational,heinzerling-inui-2021-language,10.1162/tacl_a_00324}. Thus, they are perfect backbone models for text generation and knowledge-intensive downstream tasks, such as question answering (QA).
Despite their success, these models are prone to generate text that is factually incorrect or inconsistent with a provided instruction or knowledge source; such generations are usually referred to as \emph{hallucinations} \cite{ji2023survey,zhang2023siren,bang2023multitask}. 

%This behaviour is a real concern as LLMs are increasingly becoming an integral part of modern AI systems.

\begin{figure*}[t]
    \centering
    \includegraphics[width=1.0\linewidth]{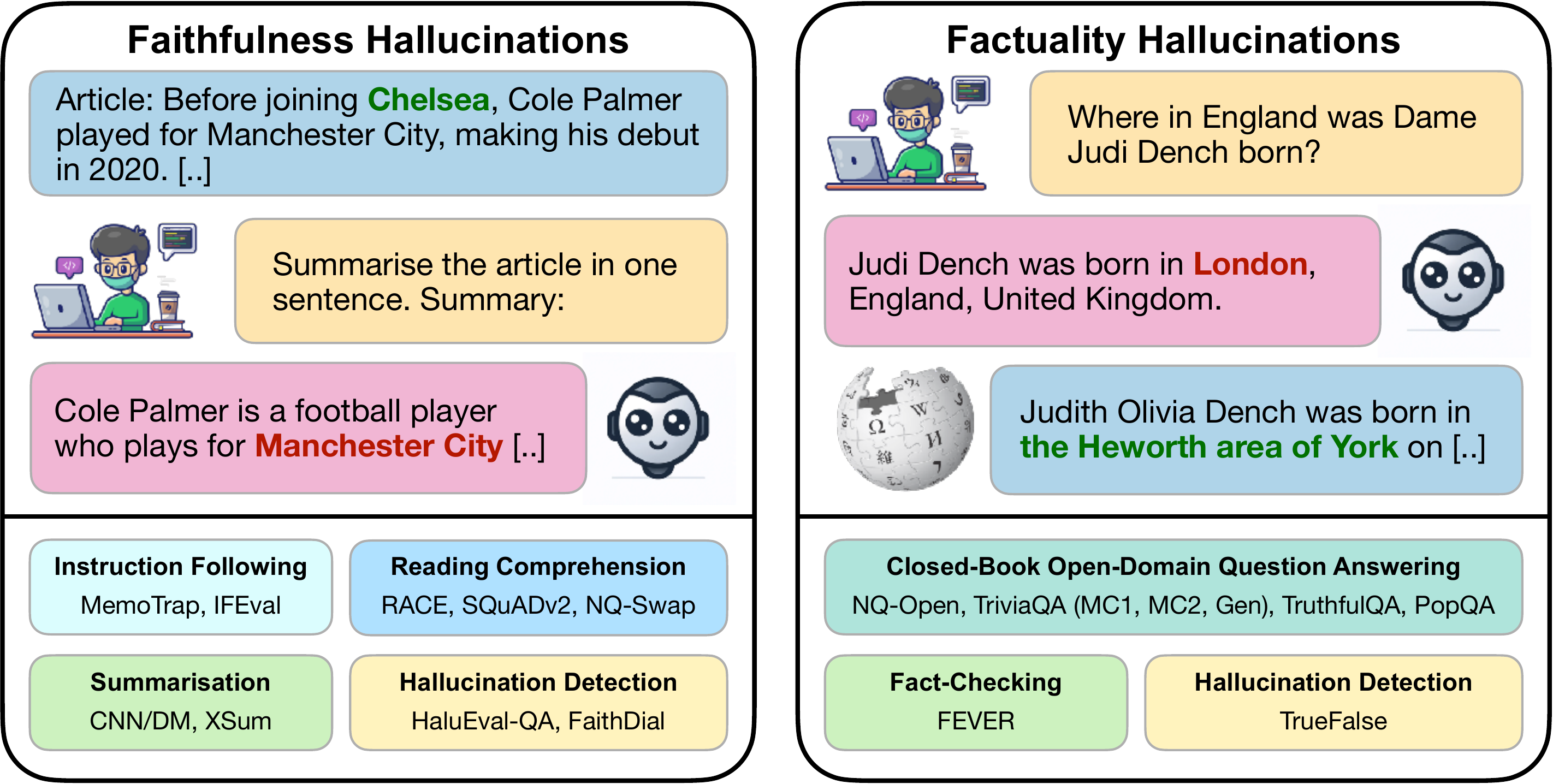}
    
    \caption{Example of LLM factuality hallucination and factuality evaluation tasks in the Hallucination Leaderboard (left). Faithfulness hallucination example and tasks on the right. 
    %\lp{very minor detail.... could we switch the order of factual/faithful boxes? also would be possible to remove the word 'Hallucination'?}
    }
    \label{fig:figure1}
\end{figure*}

In recent years, an overwhelming number of LLMs have been made available. They differ in the training approach (language modelling, instruction following, human feedback), training data used, and the number of parameters. Given the large-scale setting (i.e., number of models, their size and number of downstream tasks), it becomes difficult to gauge performance differences amongst LLMs. To systematically quantify the impact of hallucinations in several downstream tasks, we present the \emph{Hallucinations Leaderboard}\footnote{Available at \url{https://huggingface.co/spaces/hallucinations-leaderboard/leaderboard}}, a platform for evaluating the hallucination tendencies of LLMs.

We aim to reveal the hallucination tendencies of LLMs in their role as backbone models on different generative and knowledge-intensive tasks. We distinguish two scenarios for LLM hallucinations~\citep{DBLP:journals/corr/abs-2311-05232}. 
One is related to \emph{faithfulness}, i.e., whether an LLM generation adheres to the given source of information (\eg when summarising a document). The other is related to \emph{factuality}, \ie whether LLMs generate factually correct content according to world knowledge based on knowledge acquired during training (e.g., in closed-book general domain QA tasks).  
\autoref{fig:figure1} (left-top) shows an example of faithfulness hallucination where the generated summary contradicts the input document; and an example of factuality hallucination (right-top) in question answering where the model answers that \emph{Charles Lindbergh} was the first person to walk on the moon.
Concretely, we use a set of tasks, listed in \autoref{fig:figure1} (bottom), to assess LLMs' hallucination behaviour in terms of factuality and faithfulness. We evaluate 20 LLMs across 15 tasks, and each model is evaluated with no training in a zero- or very few-shot in-context examples.

Our results show variances across models and tasks, offering insights into the strengths and weaknesses of different LLMs in handling hallucinations. These results are critical for understanding the current capabilities and limitations of LLMs in various applications.
The Hallucinations Leaderboard represents a significant step towards addressing the challenge of hallucinations in LLMs. It will not only aid researchers and engineers in selecting more reliable models but also drive the development of LLMs.
The project welcomes contributions and feedback, indicating its evolving nature and commitment to continuous improvement.

% \begin{figure}[ht]
%     \centering
%     \begin{subfigure}{0.45\textwidth}
%         \centering
%         \includegraphics[width=\textwidth]{latex/figures/compare_instruct.pdf}
%         \caption{First subfigure}
%         \label{fig:sub1}
%     \end{subfigure}\hfill
%     \begin{subfigure}{0.45\textwidth}
%         \centering
%         \includegraphics[width=\textwidth]{latex/figures/compare_model_size.pdf}
%         \caption{Second subfigure}
%         \label{fig:sub2}
%     \end{subfigure}
%     \caption{Two parallel figures}
%     \label{fig:test}
% \end{figure}

\section{Evaluation Framework}
%\label{sec:overview}
\label{sec:experiment_settings}

The Hallucinations Leaderboard leverages the EleutherAI Language Model Evaluation Harness~\cite{eval-harness}, a framework for zero-shot and few-shot language model evaluation via in-context learning on a wide array of tasks.
The leaderboard covers a range of tasks, including Closed-book Open-domain QA, Summarisation, Reading Comprehension, Instruction Following, Fact-Checking, Hallucination Detection, and Self-Consistency.
Each task is designed to target specific aspects of hallucination in LLMs.
%
%
%\lp{say something else about the leaderboard interface? screen-shot?}
%
%
%Our results show variances across models and tasks, offering insights into the strengths and weaknesses of different LLMs in handling hallucinations. These results are critical for understanding the current capabilities and limitations of LLMs in various applications.

%\section{Metrics, Tasks and Models}
%\du{Rewrite and refactor datasets and metrics section}
%
%The Leaderboard consists of diverse tasks used to assess the tendency of LLMs to generate hallucinated content.
%
These tasks are generally categorised into two classes based on the type of hallucinations the models may generate: factuality hallucination and faithfulness hallucination. 
\subsection{Factuality Evaluation}
% NQ-open, TriviaQA, PopQA, FEVER, TrueFalse

\paragraph{Closed-book Question Answering}
This category involves evaluating the LLM's ability to answer questions without external knowledge sources. Natural Questions \cite{kwiatkowski-etal-2019-natural, lee-etal-2019-latent} and TriviaQA \cite{2017arXivtriviaqa} demand the generation of answers to real-world or trivia questions, assessed against the gold standard answers. PopQA \cite{DBLP:conf/acl/MallenAZDKH23} poses a new challenge by introducing questions about long-tail entities, which enables a fine-grained analysis of LLM's memorisation of factual knowledge. In addition, we measure the ability of LLMs to answer questions about the truthfulness of a statement on TruthfulQA \cite{lin2022truthfulqa}. Models are evaluated by accuracy on the multi-label classification task (MC2) in TruthfulQA.
% in 8-shot and 64-shot settings.

\paragraph{Fact-Checking} These tasks evaluate the LLM's ability to verify the authenticity of statements. Each instance in FEVER \cite{thorne-etal-2018-fever} comprises a claim and a label (\texttt{SUPPORTS} and \texttt{REFUTES}), and the model's task is to predict the label based on the claim, akin to a closed-book open-domain QA setting. The evaluation is conducted in a 16-shot setting, emphasising the model's discernment and verification capabilities.

\paragraph{Hallucination Detection}
True-False \cite{azaria-mitchell-2023-internal} assesses the model's ability to distinguish between factual and false statements across various domains. We measure the performance of LLMs by accuracy. 

\subsection{Faithfulness Evaluation}
% Summarisation: CNN/DM [x], XSum [x]
% MemoTrap [x], IFEval [x], RACE [x], SQuaD-v2 [x],  HaluEval-QA [x], Faithdial [x], TruthfulQA-mc2 [x], NQ-swap [x]
\paragraph{Summarisation} Summarisation tasks test the LLM's capability to generate concise summaries that faithfully reflect the information in the input article. XSum \cite{Narayan2018DontGM} targets single-sentence summarisation of news articles, while CNN/DM  (CNN/Daily Mail; \citealp{see-etal-2017-get}) involves generating multi-sentence summaries of news articles. Models are evaluated based on 
ROUGE-L \cite{lin-2004-rouge}, which assesses n-gram overlap with reference summaries.
% factKB (evaluating factual accuracy), and BERTScore-Precision (measuring semantic similarity between generated summary and the reference).
% all within a 2-shot learning framework. 
%\lp{Other datasets rather than CNN/DM and XSum, which are considered kind of solved w/ LLMs, should be considered. Btw ROUGE in fig 4 look rather low, do you know why? Can we add WikiCatSum and arxiv or GovReport?}

\paragraph{Reading Comprehension} These tasks examine the LLM's proficiency in understanding and extracting information from given passages. RACE \cite{lai-etal-2017-race} entails answering questions from English exam passages, while SQuAD 2.0 \cite{rajpurkar2018know} contains answerable and unanswerable questions about Wikipedia articles, requiring the LLM to discern when provided information is insufficient or ambiguous. A faithful LLM should be able to identify unanswerable questions and refuse to provide fabricated answers. Models are evaluated using Exact Match (EM) on SQuAD-v2 and by accuracy on RACE, respectively. 
% The performance is evaluated in 2-shot and 4-shot settings, respectively, emphasizing the model's interpretative and inferential abilities.

A main cause of faithfulness hallucination is that LLMs tend to rely on the memorisation of training data.
We measure the tendency of LLMs to rely on parametric knowledge using NQ-Swap~\cite{longpre-etal-2021-entity}, a dataset derived from Natural Questions \cite{kwiatkowski-etal-2019-natural, lee-etal-2019-latent}, where the gold answer in the input document is replaced by a random entity of the same entity type.
A faithful model is required to generate the replaced answer given the perturbed context. Models are evaluated by exact match based on substituted answer entities.  

\paragraph{Instruction Following} Faithful LLMs are expected to follow instructions provided by the user. We assess the LLM's fidelity in adhering to specific instructions by the following tasks. MemoTrap \cite{liu2023memotrap} involves completing text, translation, or answering questions without relying on memorised text or concepts, gauging the model's creative adherence to the given prompts in a zero-shot setting. IFEval \cite{zhou2023instruction} presents a more complex challenge, requiring the execution of a set of detailed instructions, testing the model's compliance and accuracy in following multi-faceted directives in a zero-shot evaluation. We measure the performance of LLMs on these tasks by accuracy. 

\paragraph{Hallucination Detection} These tasks are explicitly designed to detect hallucinations in LLM-generated content. FaithDial \cite{dziri2022faithdial} focuses on detecting faithfulness in dialogues. HaluEval \cite{li-etal-2023-halueval} extends this to QA, dialogue, and summarisation tasks, requiring models to identify hallucinated content in responses based on given knowledge snippets. In the leaderboard, we only consider the QA task from HaluEval, which contains human-annotated hallucinated samples created from HotpotQA \cite{DBLP:conf/emnlp/Yang0ZBCSM18}. Models are evaluated by accuracy on these tasks. 
% These tasks are performed in 8-shot or zero-shot settings, highlighting the model's ability to maintain factual integrity and coherence.

% \paragraph{Open-book Question Answering}
% \du{Refactor this section}
% These tasks examine the LLM's ability to answer a question based on the provided context. A main cause of faithfulness hallucination is that LLMs tend to rely on the memorisation of training data. We measure the tendency of LLMs to memorise by NQ-Swap~\cite{longpre-etal-2021-entity}, where the gold answer entity in the input document is substituted by a random entity. A faithful model is required to generate the replaced answer given the perturbed context. In addition, we measure the ability of LLMs to answer questions about the truthfulness of a statement on TruthfulQA \cite{lin2022truthfulqa}. Models are evaluated by exact match on NQ-Swap and evaluated by accuracy on the multi-label classification task (MC2) in TruthfulQA.

\subsection{Overall Evaluation Metrics}
\label{subsec:evaluation_metrics}
We propose two scores, namely the \emph{factuality score} and the \emph{faithfulness score}, to measure the overall performance of LLMs on each type of hallucination.
The scores are computed by averaging all the evaluation metrics on each category of tasks.

\subsection{Large Language Models}
\label{subsec:large_language_models}
%\gw{Should we list all the models? Or just the model types (pre-trained/fine-tuned/instruction-tuned/RL-tuned)?, Also, some of the models are not mentioned in our analysis. Should we keep them in our figures/results?}
The leaderboard encompasses open-source LLMs of various sizes, 
%spanning from 125 million to 40 billion parameters, 
and categorised into pre-trained, fine-tuned, and instruction-tuned models\footnote{For our analyses, we selected only a subset of models and tasks; however, the hallucination leaderboard encompasses a wider array of task metrics and models}. 

\paragraph{Base Models}
\emph{Base models} refers to LLMs that have been pre-trained on a large dataset.
We selected multiple variants of pre-trained models of different sizes:
GPT-J-6B \cite{gpt-j}, GPT-Neo 125M/1.3B/2.7B \cite{gpt-neo}, 
Bloom-560M/1.7B/7.1B \cite{workshop2022bloom}, Llama-2-7B/13B \cite{touvron2023llama}, 
Mistral-7B \cite{jiang2023mistral}, and Falcon-7B \cite{almazrouei2023falcon}.

%\paragraph{Fine-tuned Models}
%
%``Fine-tuned models'' are pre-trained models that have been further fine-tuned on a smaller, more specific dataset.
%We compare the fine-tuned models with their corresponding base pre-trained models to analyse the impact of fine-tuning on the LLMs' tendency to hallucinate.
%vicuna-class-shishya-7b-ep3/vicuna-class-tutor-7b-ep3 \cite{sonkar2023class}, Wizard-Vicuna-7B-Uncensored-GPTQ/Wizard-Vicuna-13B-Uncensored-GPTQ, 
%We select zephyr-7b-beta \cite{tunstall2023zephyr} and OpenHermes-2.5-Mistral-7B that are fine-tuned variants of Mistral-7b, as well as Orca-2-13b \cite{zheng2024judging} that is a fine-tuned variant of Llama2-13b.

\paragraph{Fine-tuned Models}
\emph{Fine-tuned models} are pre-trained models that have been further fine-tuned on a specific dataset and task to improve certain capabilities.
One variation of fine-tuning techniques is \emph{instruction fine-tuning}, which further fine-tunes a base model on a dataset of instructions\footnote{A chat or instructions dataset}, aiming to enhance their ability to follow human directives.
We selected several instruction-tuned models such as Llama-2-7b/13b-chat \cite{touvron2023llama} and Vicuna-7b-v1.5 \cite{zheng2023judging}, which are instruction-tuned versions of Llama-2 models.
Falcon-7b-instruct \cite{almazrouei2023falcon} and Mistral-7b-instruct \cite{jiang2023mistral} are instruction-tuned versions of Falcon-7b and Mistral-7b, respectively.
Another fine-tuning technique is Reinforcement Learning with Human Feedback (RLHF), for example, via Direct Preference Optimisation~\citep[DPO,][]{DBLP:conf/nips/RafailovSMMEF23}. We selected zephyr-7b-beta \cite{tunstall2023zephyr} as a representative model that is fine-tuned via RLHF.
% SOLAR-10.7B-Instruct-v1.0 \cite{kim2023solar}

The leaderboard aims to analyse the effect of scale and type on the LLMs' tendency to hallucinate.
% For simplicity, we choose representative ones per scale and type
For simplicity, we undertake experiments and analyses solely on a selected few representative models from each scale and type.

\begin{figure*}[th!]
    \centering
    \includegraphics[scale=0.31]{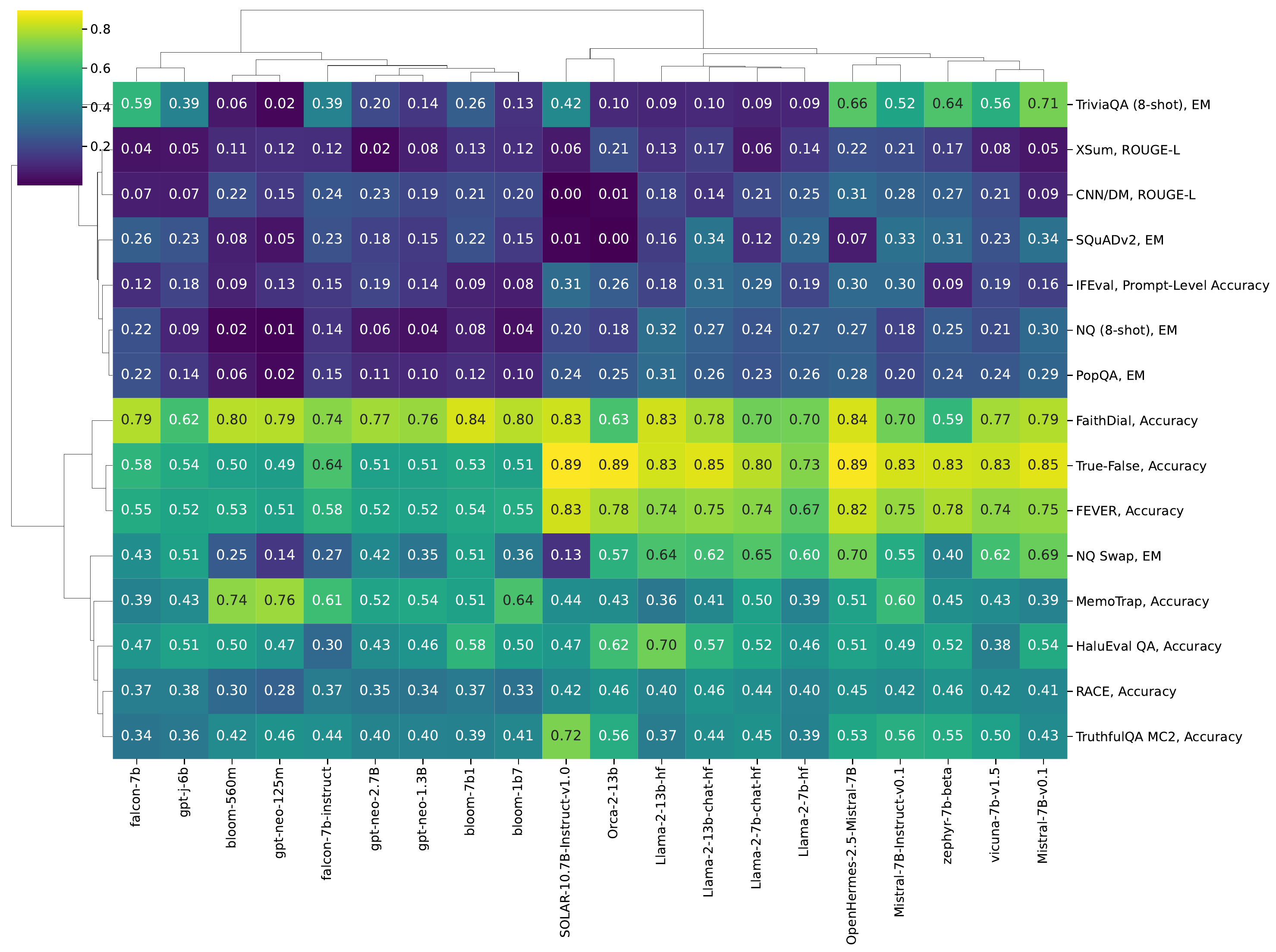}
    \caption{The heatmap of results for various tasks of selected models. Each value in the heatmap follows the corresponding task's metric while clustering on the $x$-axis and $y$-axis was done after model/task normalisation. %\lp{For summarisation tasks, factKB score should be reported instead of ROUGE}
    }
    \label{fig:heatmap_all}
    %\vspace{-0.3cm}
\end{figure*}

\begin{figure}[th]
\centering
\subfigure{
\centering
\begin{minipage}[t]{1\linewidth}
    \centering
    \includegraphics[width=0.95\linewidth]{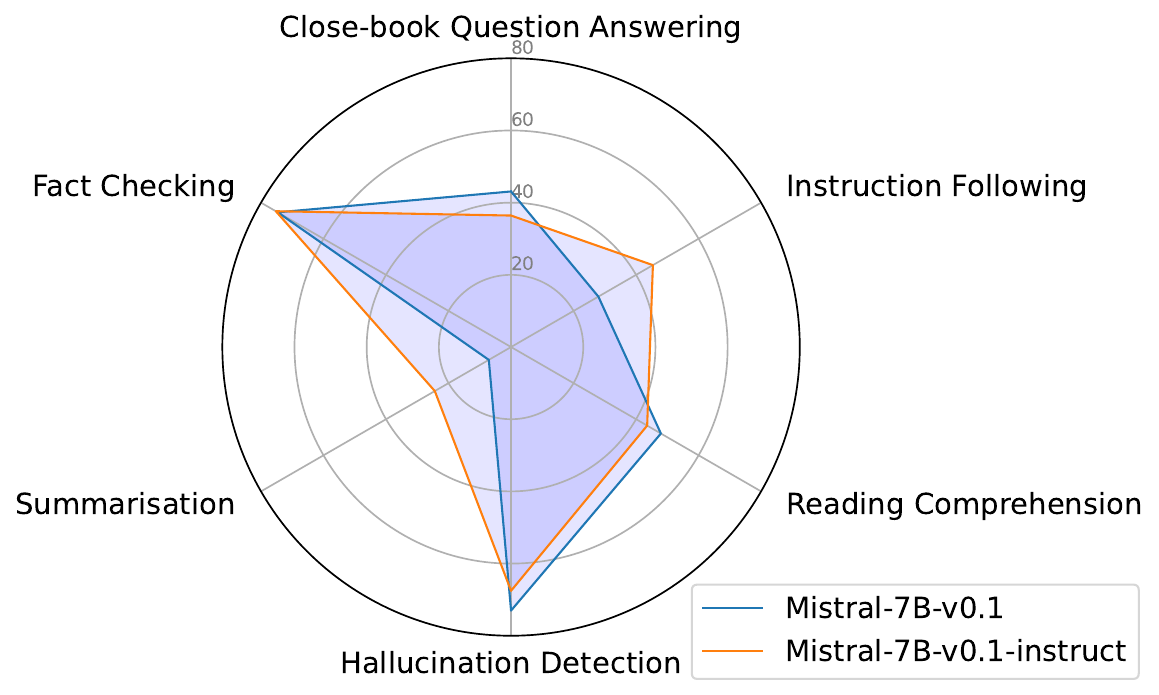}
    \caption{Comparison of models with and without instruction fine-tuning.}
    \label{fig:radar-instruct}
\end{minipage}
} \\
\subfigure{
\centering
\begin{minipage}[t]{0.95\linewidth}
    \centering
    \includegraphics[width=1\linewidth]{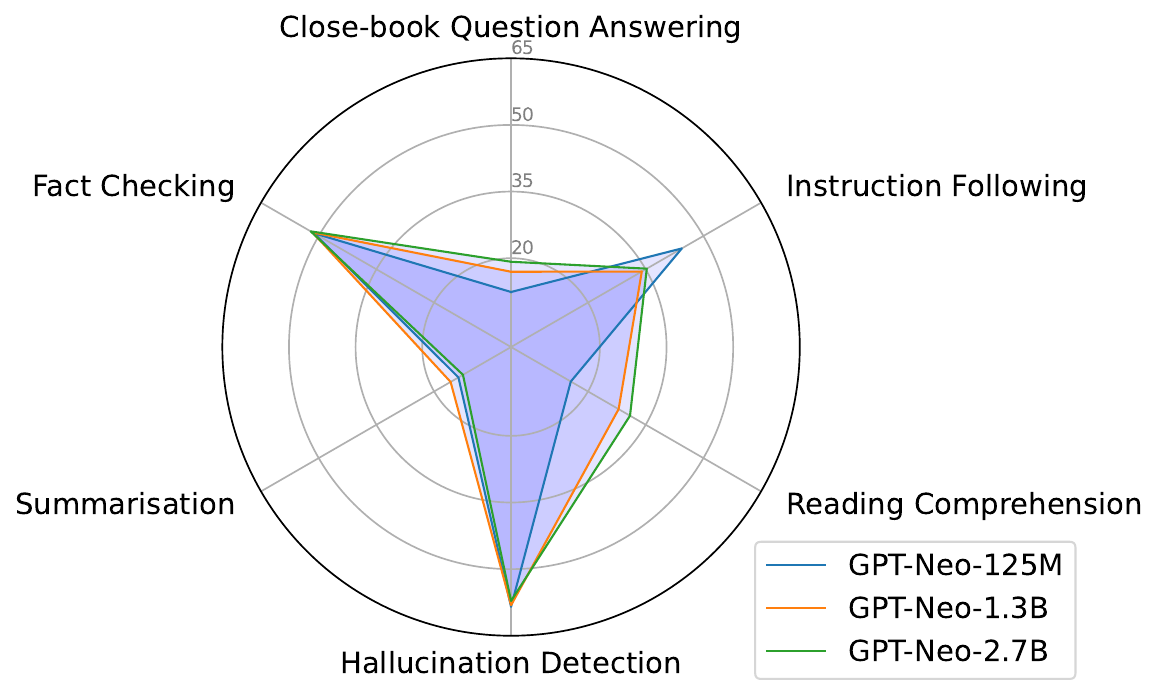}
    \caption{Comparison of models with different sizes.}
    \label{fig:radar-size}
    \end{minipage}
}%
%\vspace{-0.2cm}
\end{figure}

%\section{Comprehensive Analysis of Hallucination in LLMs}
\section{Results}
\label{sec:comprehensive_alaysis}

To gain a deeper understanding of hallucinations in LLMs, we conducted a comprehensive analysis of the models and tasks introduced in Section \ref{sec:experiment_settings}. In \autoref{fig:heatmap_all}, we display the results of models for each task in the form of a heatmap. The value of each cell in the heatmap follows the metric of the corresponding task, and the dendrogram-shaped clusters are formed after applying min-max normalisation by task (y-axis) and model (x-axis). 
The hierarchical clustering is computed using the Ward variance minimisation linkage method~\cite{ward1963hierarchical} and Euclidean distance to group similar data points based on their mean pairwise distances, organising them into a tree structure.
%How Difficult are Certain Tasks for LLMs?

\autoref{fig:heatmap_all} shows the task-related hallucination tendency of LLMs.
We observe that LLMs are better at judging factuality and faithfulness than what they are at producing factual and faithful generations. Llama-2 models \cite{touvron2023llama} show the stronger opposite behaviour, i.e., they perform relatively well in FEVER, FaithDial and true-false while quite poorly in QA tasks such as NQ-open. Mistral models perform slightly better on the TriviaQA task. This agrees with the findings in \citet{li2023inferencetime} and~\citet{Zhang2023HowLM} where authors find that models have better internal representations of truthfulness than what they often surface.
Second, tasks that mostly require completing a text sequence (\eg, MemoTrap or TruthfulQA-MC2) reflect slightly better performance than those that involve reading a longer input context (\eg XSum) or answering a question based on memorised knowledge (\eg NQ-open). 

By examining how models are clustered, it becomes evident that the hallucination tendency is less dependent on the model type (Section \ref{subsec:large_language_models}) and more on their belonging to the same family --- \eg Llama-2 \cite{touvron2023llama}, GPT-Neo \cite{gpt-neo}, Bloom \cite{workshop2022bloom}. This observation can be attributed to the fact that while models from different families may possess distinct training data and structures, those within the same family generally share architectures and are based on the same pre-training data. This finding has inspired us to analyse the impact of instruction fine-tuning and the influence of model size within the same family, as elaborated in Sections \ref{subsec:impact_chat} and \ref{subsec:model_size}.

\subsection{Impact of Instruction Fine-Tuning on Hallucinations}
\label{subsec:impact_chat}

\begin{table}[t!]
% \small
\centering
\scalebox{0.78}{
\begin{tabular}{lcc}
\toprule
\textbf{Models}          & \textbf{Faithfulness} & \textbf{Factuality} \\
\midrule
Llama-2-7b               & 37.94 (+0.0)  &40.12 (+0.0)  \\
Llama-2-7b-Chat          & 38.69 (\textcolor{lightblue}{+0.8})  & 42.48 (\textcolor{lightblue}{+2.4})  \\
Vicuna-7b-v1.5          & 37.13 (\textcolor{red}{-0.8})  & 51.42 (\textcolor{lightblue}{+11.3})  \\
\midrule
Llama-2-13b              & 39.75 (+0.0)  &44.49 (+0.0)  \\
Llama-2-13b-Chat         & 42.32 (\textcolor{lightblue}{+2.6})  & 44.60 (\textcolor{lightblue}{+0.1})  \\
\midrule
Mistral-7B-v0.1          & 38.62 (+0.0)  & 55.41 (+0.0)  \\
Mistral-7B-Instruct-v0.1 & 43.26 (\textcolor{lightblue}{+4.6})  & 50.74 (\textcolor{red}{-4.7}) \\
Zephyr-7b-beta & 36.14 (\textcolor{red}{-2.5})  & 55.11 (\textcolor{red}{-0.3}) \\
OpenHermes-2.5-Mistral-7B & 43.88 (\textcolor{lightblue}{+5.3})  & 57.41 (\textcolor{lightblue}{+2.0}) \\
\midrule
Falcon-7b                & 32.81 (+0.0)  & 41.74 (+0.0)  \\
Falcon-7b-Instruct       & 33.61 (\textcolor{lightblue}{+0.8})  & 38.90 (\textcolor{red}{-2.8})  \\
\bottomrule
\end{tabular}
}
\caption{Comparison between the faithfulness and factuality scores (introduced in Section \ref{subsec:evaluation_metrics}) produced by the base models and their corresponding fine-tuned models. 
Performance differences are against the base models.
%; the 
%$\Delta$ is against the pre-trained model.
 % \textbf{Bold cell} indicates the best-performing model variant for a given metric.
}
\label{table:pre-trained_vs_instruction_following}
\end{table}
\autoref{table:pre-trained_vs_instruction_following} shows a comparison of pre-trained models with their corresponding instruction fine-tuned variants across two metrics: \emph{faithfulness score} and \emph{factuality score} (Section \ref{subsec:evaluation_metrics}). We can observe that instruction fine-tuned models achieve higher Faithfulness scores than their base counterparts. This is indicative of their enhanced ability to retain fidelity to the given input or the specific instructions given (\eg \emph{"Answer the following question based on the provided context"}). 

Unlike Faithfulness, Factuality scores across the models show a trend of either marginal improvement or, in some cases, a decline, with the notable exception of the Llama-2-7b model. While these models become better at adhering to instructions or the input, which is reflected in the improved Faithfulness scores, their capacity to produce factually accurate information does not consistently improve in the same way. 
This pattern suggests a trade-off between Faithfulness and Factuality in instruction fine-tuning: enhancing a model's ability to follow instructions closely (Faithfulness) might not always lead to improvements in the accuracy of the information produced (Factuality).

To analyse the impact of instruction fine-tuning on hallucinations further, in \autoref{fig:radar-instruct}, we compare Mistral-7B models \cite{jiang2023mistral} with and without instruction fine-tuning in different categories of tasks defined in Section~\ref{sec:experiment_settings}.
We can see that the improvement in faithfulness is mainly attributed to the improvement in instruction fine-tuning and summarisation tasks, while the decrease in factuality is caused by the degradation of question-answering and hallucination-detection tasks.

\begin{table}[t!]
\small
\centering
\begin{tabular}{lcc}
\toprule
\textbf{Models} & \textbf{Faithfulness} & \textbf{Factuality} \\
\midrule
%gpt-neo         &                       &                     \\
GPT-Neo-125m          &32.08 (+0.0) &25.04 (+0.0) \\
GPT-Neo-1.3B          &33.91 (\textcolor{lightblue}{+1.8})  &28.36 (\textcolor{lightblue}{+3.3})  \\
GPT-Neo-2.7B          &34.28 (\textcolor{lightblue}{+2.2})  &29.91 (\textcolor{lightblue}{+4.9}) \\
\midrule
%bloom           & \textbf{}             &                     \\
Bloom-560m          &34.32 (+0.0) &26.52 (+0.0) \\
Bloom-1b7           &35.18 (\textcolor{lightblue}{+0.9}) &28.80 (\textcolor{lightblue}{+2.3}) \\
Bloom-7b1           &38.38 (\textcolor{lightblue}{+4.1}) &32.07 (\textcolor{lightblue}{+5.6}) \\
\midrule
%Llama-2         &                       &                     \\
Llama-2-7b             &37.94 (+0.0) &40.12 (+0.0) \\
Llama-2-13b            &39.75 (\textcolor{lightblue}{+1.8}) &44.49 (\textcolor{lightblue}{+4.4}) \\
\midrule
%Llama-2-chat    &                       &                     \\
Llama-2-chat-7b        &38.69 (+0.0) &42.48 (+0.0) \\
Llama-2-chat-13b       &42.32 (\textcolor{lightblue}{+3.6}) &44.60 (\textcolor{lightblue}{+2.1})\\
%\midrule
%falcon          &                       &                     \\
% Falcon-7b             &32.81 ($\Delta${+0.0}) &41.74 ($\Delta${+0.0}) \\
% Falcon-40b            &-  &-  \\
% \midrule
%falcon-instruct &                       &                     \\
% Falcon-instruct-7b    &33.61 ($\Delta${+0.0}) &38.90 ($\Delta${+0.0}) \\
% Falcon-instruct-40b   &-  &-  \\
\bottomrule
\end{tabular}
\caption{Comparison between the faithfulness and factuality scores produced by models of different scales, where differences are against the smallest models.}
\label{table:model_size}
\end{table}

\subsection{Impact of Model Size on Hallucinations}
\label{subsec:model_size}

To explore the impact of model size on Faithfulness and Factuality hallucinations, we provide Faithfulness and Factuality scores across various model sizes in \autoref{table:model_size}. While an increase in model size generally enhances both Faithfulness and Factuality, it is noteworthy that Factuality tends to exhibit more substantial improvements compared to Faithfulness, with the Llama-2-chat models being a notable exception to this trend. 

This suggests that as model size increases, the accumulation of parametric knowledge during the training phrase becomes more extensive, leading to a reduced dependence on context. This observation aligns with the findings of existing research, such as "Imitative Falsehoods" \citep{lin2022truthfulqa} or "Strong Prior" \citep{mckenzie2022inverse}, expanding upon them across various models and tasks.

In \autoref{fig:radar-size}, we show the evaluation results of different sizes of GPT-Neo on different categories of tasks.
We observe that GPT-Neo can obtain higher accuracy on question-answering and open-book question-answering tasks when the model's size is increased, contributing to improved factuality.
We can also see that GPT-Neo-125M is more accurate on instruction-following tasks than larger models, which is mainly due to the "Strong Prior" phenomenon, as we discussed above.
%
% This is mainly because MemoTrap is specifically designed for studying inverse scaling patterns in LLMs.
%
% Our observation matches the finding of inverse scaling \cite{mckenzie2022inverse}, which states that larger models tend to rely on memorisation of training data rather than following instructions.
%

\section{Related Work}

%\gw{please add existing studies on hallucination}

%Previous studies have investigated hallucinations in LLMs.
Much work has focused on hallucination detection for summarisation tasks (\citealt{maynez-etal-2020-faithfulness,kryscinski-etal-2020-evaluating,scialom-etal-2021-questeval,ribeiro-etal-2022-factgraph,laban-etal-2022-summac,utama-etal-2022-falsesum,sentli}). 
For instance, \citet{laban-etal-2022-summac} propose SummaC, which examines faithfulness through a Natural Language Inference framework. 
Far from solved, this problem takes a broader scope in the context of LLMs \cite{DBLP:journals/corr/abs-2311-05232,ye2023cognitive}.
Factual and faithfulness evaluations are carried out for LLMs' diverse downstream tasks and by LLM evaluators. For instance, \citet{chen2023felm} propose a benchmark covering question answering, reasoning, maths, and writing recommendation tasks. \citet{chuang2024dola} propose a decoding strategy to improve factuality on multiple choice and open-ended generation tasks.
Some work proposes LLM-based hallucination evaluators \cite{cohen-etal-2023-lm,zhang-etal-2023-sac3,manakul2023selfcheckgpt}. For instance, \citet{cohen-etal-2023-lm} propose a multi-turn iterative examination between LLMs where one LLM formulates claims and the other asks questions to uncover inconsistencies. Our Hallucination Leaderboard supports this research gathering for evaluation LLMs and downstream tasks.

%However, much of the work has concentrated on specific tasks of dimensions of performance.

% DecodingTrust is a robust study on the trustworthiness of LLM~\cite{wang2023decodingtrust}. They focus on different metrics of trustworthiness, including toxicity, stereotypical bias, and robustness against adversarial and out-of-distribution samples. We focus on the lower-level aspects of hallucination which are faithfulness and factuality.
Leaderboards have arisen in 2023 as a way to quickly get insights on model capabilities, by comparing models in equivalent and reproducible setups \cite{open-llm-leaderboard,wang2023decodingtrust}. They contribute significantly to our understanding of LLMs' capabilities and limitations in specific areas.
Looking at the same domain, the Hughes Hallucination Evaluation Model (HHEM) leaderboard \citep{HughesBae2023} focuses on a summarisation tasks, and uses a model as a judge approach to evaluate hallucinations.
Our study and leaderboard aim to broaden the scope by evaluating hallucinations in LLMs across-the-board, using a wide variety of tasks and metrics. We aim to complement and extend the insights gained from existing studies, providing a more comprehensive understanding of LLMs' strengths and weaknesses in terms of hallucinations.
% HELM?
% 
% Note Galileo doesn't have a citable resource

%\section{Future Directions}
%\label{sec:future_directions}

%The Hallucinations Leaderboard represents a significant step towards addressing the challenge of hallucinations in LLMs. It not only aids researchers and engineers in selecting more reliable models but also drives the development of LLMs for more accurate and faithful language generation. The project welcomes contributions and feedback, indicating its evolving nature and commitment to continuous improvement.

\section{Conclusions}
\label{sec:conclusions}

The Hallucinations Leaderboard provides a platform for understanding and mitigating hallucinations in LLMs.
By offering a comprehensive evaluation across a diverse set of benchmarks, it enables a deeper understanding of the generalisation properties and limitations of large language models.
This initiative marks a pivotal step towards enhancing the reliability and effectiveness of LLMs in real-world settings.

\paragraph*{Acknowledgements}
Experiments are being conducted mainly at the Edinburgh International Data Facility (EIDF) and on the internal clusters of the School of Informatics, University of Edinburgh.
APG was supported by the United Kingdom Research and Innovation (grant EP/S02431X/1), UKRI Centre for Doctoral Training in Biomedical AI at the University of Edinburgh, School of Informatics.
PM was partially funded by ELIAI (The Edinburgh Laboratory for Integrated Artificial Intelligence), EPSRC (grant no. EP/W002876/1); an industry grant from Cisco; and a donation from Accenture LLP; and is grateful to NVIDIA for the GPU donations.
XH and PM are funded by an industry grant from Cisco.
GW was supported by ILCC program (School of Informatics Funding Package) at the University of Edinburgh, School of Informatics.
RS, XD, and YZ are supported in part by the UKRI Centre for Doctoral Training in Natural Language Processing, funded by UK Research and Innovation (grant EP/S022481/1) and the School of Informatics.

\section*{Limitations}
In this work, we define a prompt template for each task. Although we analysed the robustness of prompt templates for some tasks in Appendix \ref{appendix:template_robust}, what constitutes an appropriate template from a hallucination perspective has not been sufficiently considered for all tasks. Additionally, despite the possibility that each model may have an optimal custom prompt template, we are defining task-specific templates without considering this. Addressing this issue further is one of our future objectives.
Furthermore, the number of demonstrations (shots) for in-context examples has not been sufficiently explored. This can particularly impact tasks related to faithfulness, where obtaining necessary information from the given context is crucial; in-context learning could be utilised to encourage models to remain faithful to the provided context.
Finally, due to cost reasons, we only considered open-source models and did not take closed-source models like GPT-4~\citep{DBLP:journals/corr/abs-2303-08774} into account.

\section*{Ethics Statement}
This paper analyses the issue of hallucination in LLMs, which by itself can have a broader social impact due to the possibility of misinformation in the case of hallucinated outputs.
We aim to provide researchers and practitioners with empirical results on model hallucinations and spread awareness of this phenomenon in general, hopefully reducing the possibility of reliance on factually incorrect LLM outputs.
Furthermore, our leaderboard covers a selection of tasks proposed in prior research, and the datasets of these tasks can contain bias due to their collection protocols.
Therefore, the reported results might be affected by the lack of some demographic and societal groups from those datasets and the over representation of others~\cite{hovy2021bias}. 
We acknowledge this limitation and encourage the creators of further hallucination detection benchmarks to consider it during the data collection process.
% Lastly, large language models are known to amplify existing social biases~\cite{gallegos2023bias}

\bibliography{custom}

\appendix

\clearpage

% \section{Prompt Templates}
% \label{appendix:prompts_templates}

\section{Prompt Template Robustness}
\label{appendix:template_robust}
As shown by recent research, evaluation of LLM capabilities can produce results that are highly dependent on the exact format of the prompt, including the formulation of the instruction and the template for few-shot demonstrations~\cite{sclar2023quantifying,mizrahi2023state,voronov2024mind}.
To reflect these findings in the design of our study, we conduct preliminary experiments on the sensitivity of our evaluation results to minor prompt variations, intending to supplement future versions of the leaderboard with such measurements.
For simplicity and due to the computational cost of evaluation with multiple prompts, we explore a subset of models and tasks from our main results, as well as a small number of prompt variations.

More specifically, for Llama-2-7b and Llama-2-7b-chat models, we generate 5 paraphrases of instructions for the Natural Questions dataset using gpt3.5-turbo as described in~\citet{mizrahi2023state} and generate 3 variations of the prompt in the TruthfulQA MC2 dataset by changing the inter-example separators and input verbalisers as described in~\citet{voronov2024mind}.
After generating those variations, we run standard evaluation described previously and report average model performance on each task, as well as the standard deviation across prompt formats.

\begin{table}[t!]
\setlength{\tabcolsep}{3pt}
\begin{tabular}{@{}lcc@{}}
\toprule
Model           & NQ                      & TruthfulQA MC2 \\ \midrule
Llama-2-7b      & $0.27_{\pm 7\cdot 10^{-4}}$ & $0.39_{\pm 0.01}$   \\
Llama-2-7b-chat & $0.23_{\pm 2\cdot 10^{-3}}$ & $0.45_{\pm 0.01}$   \\ \bottomrule
\end{tabular}
\caption{Prompt template evaluation results for Llama-2-7b and Llama-2-7b-chat. Standard deviations across 5 instructions for Natural Questions and 3 prompts for TrtuthfulQA are given in the subscript.}
\label{tab:template-robustness}
\end{table}

The results of this experiment can be found in Table~\ref{tab:template-robustness}.
Notably, while introducing prompt variations leads to changes in the evaluation results, the changes themselves are relatively minor.
There are two possible explanations to this phenomenon: first, the set of prompts which we use for evaluation is more narrow compared to prior work, and additional generated instructions could lead to more noticeable distortions in task performance.
Second, both tasks we use for evaluation rely on factual knowledge of the models and the ability to extract and analyse factual information from inputs.
Most prior work on prompt robustness has used tasks that are more dependent on logical reasoning or understanding of surface-level linguistic features, which might be more sensitive to changes in the formulations of the prompt.

\section{Factuality in summarisation Tasks}

Table~\ref{table:factKB} shows FactKB scores \cite{feng-etal-2023-factkb} for summarisation tasks. An overall observation is that scores are relatively high, specially for CNN/DM, indicating that models seem to generate factual content. However, this contrast with the low ROUGE-L scores in Figure~\ref{fig:heatmap_all}. We speculate that models are generating related factual content but which is potentially not salient (i.e., fail to do the abstractive summarisation task). When confronting different training regimes, i.e., pre-trained vs instruction-tuned, we see differences across model families in the more abstractive XSum task. For Llama models, instruction-tuning seems to encourage model generations that are less factual while for Mistral and Falcon families, factuality increases with instruction-tuning.
We also observed that the FactKB classified generations from Llama 13b as less factual compared to those by the smaller 7b model. As this trend does not aligns with other results, we did not include FactKB in the main results and reported ROUGE-L instead. We plan to incorporate in the leaderboard additional factuality and faithfulness metrics for summarisation tasks \cite{scialom-etal-2021-questeval,laban-etal-2022-summac}.

\begin{table}[t!]
% \small
\centering
\scalebox{0.78}{
\begin{tabular}{lcc}
\toprule
\textbf{Models}          & \textbf{XSum} & \textbf{CNN/DM} \\
\midrule
Llama-2-7b               & 80.75 (+0.0)  & 89.77 (+0.0)  \\
Llama-2-7b-Chat          & 47.64 (\textcolor{red}{-33.11})  & 84.93 (\textcolor{red}{-4.84})  \\
\midrule
Llama-2-13b              & 77.28 (+0.0)  & 96.32 (+0.0)  \\
Llama-2-13b-Chat         & 49.17 (\textcolor{red}{-28.11})  & 91.38 (\textcolor{red}{-4.94})  \\
\midrule
Mistral-7B-v0.1          & 31.56 (+0.0)  & 95.55 (+0.0)  \\
Mistral-7B-Instruct-v0.1 & 49.26 (\textcolor{lightblue}{+17.7})  & 98.62 (\textcolor{lightblue}{+3.07})  \\
\midrule
Falcon-7b                & 39.55 (+0.0)  & 95.59 (+0.0)  \\
Falcon-7b-Instruct       & 67.47 (\textcolor{lightblue}{+27.92})  & 94.91 (\textcolor{red}{-0.68})  \\
\bottomrule
\end{tabular}
}
\caption{Results for FactKB assessing factual accuracy for summarisation tasks comparing pre-trained models and corresponding instruction fine-tuned models. $\Delta$ is against the pre-trained model.
 % \textbf{Bold cell} indicates the best-performing model variant for a given metric.
}
\label{table:factKB}
\end{table}

\begin{table*}[h!]
\begin{center}
\resizebox{\textwidth}{!}{
\begin{tabular}{ll}

Task: Natural Questions & Task: TruthfulQA \\
\toprule
%\tiny{\texttt{\textit{\underline{Question}}}}
\texttt{\underline{Question}} & \texttt{\underline{Question}} \\
\texttt{which state is located in the centre of india} \quad & \texttt{What happens if you smash a mirror} \\
&\\
\texttt{\underline{Answer}} & \texttt{\underline{Answer}}\\
\texttt{\color{lightblue}{Madhya Pradesh}} & \texttt{If you smash a mirror, \color{lightblue}{the mirror breaks.}}\\
&\\
\texttt{\underline{Bloom-560m}} & \texttt{\underline{Llama-2-7b}}\\
\texttt{\color{red}{India}} & \texttt{If you smash a mirror, \color{red}{you will get seven}}\\
& \texttt{\color{red}{years of bad luck.}}\\
\texttt{\underline{Bloom-1b7}} &\\
\texttt{\color{red}{Rajasthan}} & \texttt{\underline{Llama-2-13b}}\\
& \texttt{If you smash a mirror, \color{lightblue}{the glass will }}\\
\texttt{\underline{Bloom-7b1}} & \texttt{\color{lightblue}{break and the mirror will be destroyed.}}\\
\texttt{\color{lightblue}{Madhya Pradesh}}&\\
\bottomrule

\end{tabular}
}
\caption{Case study on the impact of model size on two Factuality tasks (Natural Questions and TruthfulQA). {\color{lightblue}{Light blue}} indicates answers that are correct or can be considered correct, while {\color{red}{red}} indicates incorrect answers.}
\label{table:case_study}
\end{center}
\end{table*}

\begin{table*}[h!]
\begin{center}
\resizebox{\textwidth}{!}{
\begin{tabular}{ll}

Task: NQ-Swap & \\
\toprule
%\tiny{\texttt{\textit{\underline{Question}}}}
\texttt{\underline{Question}} &  \\
\texttt{who plays max voice in a goofy movie} &  \\
&\\
\texttt{\underline{Original Context}} & \\
\texttt{{\color{red}{Jason Marsden}} as Maximillan "Max" Goof, Goofy 's insecure teenage son.} \quad\quad\quad\quad\quad& \\
&\\
\texttt{\underline{Substituted Context}} & \\
\texttt{{\color{lightblue}{Hugh S. Johnson}} as Maximillan "Max" Goof, Goofy 's insecure teenage son.} & \\
& \\
\texttt{\underline{Llama-2-7b's Answer}} &\\
\texttt{\color{red}{Jason Marsden}} & \\
& \\
\texttt{\underline{Llama-2-7b-chat's Answer}} & \\
\texttt{\color{lightblue}{Hugh S}}&\\
\bottomrule

\end{tabular}
}
\caption{Case study comparing the base model (Llama-2-7b) and its corresponding instruction fine-tuned version (Llama-2-7b-chat) on NQ-Swap, one of the Faithfulness Tasks. {\color{lightblue}{Light blue}} indicates the substituted answer, while {\color{red}{red}} indicates the original.}
\label{table:case_study_2}
\end{center}
\end{table*}

\section{Case Study on Generation Samples}
\label{appendix:case_stduy}
\autoref{table:case_study} shows the variation in the generated outputs according to model size for two Factuality tasks (Natural Questions, TruthfulQA). On the Natural Questions (left-hand side), for the question \textit{"which state is located in the centre of india"}, it can be seen that the smallest Bloom model \cite{workshop2022bloom}, Bloom-560m, fails to name a state and incorrectly answers \textit{"india"}. In contrast, the larger model, Bloom-1b7, shows improvement with its output \textit{"Rajasthan"}, which is indeed one of the states of India (specifically, a region adjacent to the correct answer, \textit{"Madhya Pradesh"}). However, this is still not the correct answer and exemplifies a typical case of factuality hallucination. Meanwhile, the largest model, Bloom-7b1, accurately outputs the correct answer, \textit{"Madhya Pradesh"}. In TruthfulQA (right-hand side), given the question \textit{"What happens if you smash a mirror"}, which falls under the "Superstitions" category, it's noted that the larger model, Llama-2-13b \cite{touvron2023llama}, successfully answers with the factually accurate response \textit{"the mirror breaks"}. In contrast, the smaller model, Llama-2-7b, provides an answer aligned with superstitions, stating \textit{"you will get seven years of bad luck"}.

For a case study on the Faithfulness task, in \autoref{table:case_study_2}, we compare the outputs of the base model (Llama-2-7b) and its corresponding instruction fine-tuned version (Llama-2-7b-chat) on the NQ-Swap task. We observe that the base model, without considering the changed context, retrieves the original answer \textit{"Jason Marsden"} from its parametric knowledge, indicating an instance of faithfulness hallucination. In contrast, the instruction fine-tuned model accurately reflects the changed context and generates \textit{"Hugh S"} as the correct answer. This suggests that the instruction fine-tuned model better adheres to the provided instruction, \textit{"Answer the following question based on the provided context"} (\autoref{table:additional_details}), thereby indicating it has become more faithful.

\begin{table*}[t!]
\small
\resizebox{\textwidth}{!}{
\begin{tabular}{ccccc}
\textbf{Type} &
  \textbf{Task} &
  \textbf{Metric} &
  \textbf{\# of shots} &
  \textbf{Prompt Template} \\
  \toprule
\multirow{6}{*}{\begin{tabular}[c]{@{}c@{}}Factuality\\ Hallucination\end{tabular}} &
  \begin{tabular}[c]{@{}c@{}}Natural \\ Questions\end{tabular} &
  EM &
  8 &
  "Answer these questions:\textbackslash{}n\textbackslash{}nQ: " + \{\{question\}\} + "?\textbackslash{}nA:" \\
 &
  \rule{0pt}{9pt} TriviaQA &
  EM &
  8 &
  "Question: " +\{\{question\}\} + "?\textbackslash{}nAnswer:" \\
 &
  \rule{0pt}{9pt} PopQA &
  EM &
  8 &
  "Answer these questions:\textbackslash{}n\textbackslash{}nQ: " + \{\{question\}\} + "?\textbackslash{}nA:" \\
 &
  \rule{0pt}{15pt} \begin{tabular}[c]{@{}c@{}}TruthfulQA \\ (MC2)\end{tabular} &
  Accuracy &
  6 &
  "Q: " + \{\{question\}\} + "\textbackslash{}nA:" \\
 &
  \rule{0pt}{9pt} FEVER &
  Accuracy &
  8 &
  "Claim: " + \{\{claim\}\} + '\textbackslash{}nLabel:" \\
 &
  \rule{0pt}{9pt} True-False &
  Accuracy &
  8 &
  "Statement: " + \{\{statement\}\}+ "\textbackslash{}nLabel:" \\
  \midrule\midrule
\multirow{9}{*}{\begin{tabular}[c]{@{}c@{}}Faithfulness \\ Hallucination\end{tabular}} &
  \rule{0pt}{9pt} XSum &
  ROUGE-L &
  0 &
  \begin{tabular}[c]{@{}c@{}}"Article: " + \{\{document\}\} + \\ "\textbackslash{}nSummarize the article in one sentence. Summary:"\end{tabular} \\
 &
  \rule{0pt}{15pt} CNN/DM &
  ROUGE-L &
  0 &
  "Article: " +\{\{article\}\}+ "\textbackslash{}nSummarize the article. Summary:" \\
 &
  \rule{0pt}{18pt} RACE &
  Accuracy &
  0 &
  \begin{tabular}[c]{@{}c@{}}"Article: " + \{\{article\}\} + "\textbackslash{}n\textbackslash{}n"Question: " + \{\{question\}\} + \\ "\textbackslash{}n""Answer: " + \{\{answer\_options\}\} + "\textbackslash{}n""\end{tabular} \\
 &
  \rule{0pt}{15pt} SQuADv2 &
  EM &
  4 &
  \begin{tabular}[c]{@{}c@{}}"Title: " + \{\{title\}\} + "\textbackslash{}n\textbackslash{}n" + "Background: " + \{\{context\}\} + \\ "\textbackslash{}n\textbackslash{}n" + "Question: " + \{\{question\}\} + "\textbackslash{}n\textbackslash{}n" + "Answer:"\end{tabular} \\
 &
  \rule{0pt}{21pt} NQ-Swap &
  EM &
  4 &
  \begin{tabular}[c]{@{}c@{}}"Answer the following question based on the provided \\ context:\textbackslash{}n\textbackslash{}nContext: "+ \{\{sub\_context\}\} + "\textbackslash{}nQuestion: " \\ + \{\{question\}\} + "?\textbackslash{}nAnswer:"\end{tabular} \\
 &
  \rule{0pt}{15pt} MemoTrap &
  Accuracy &
  0 &
  \{\{prompt\}\} \\
 &
  \rule{0pt}{15pt} IFEval &
  \begin{tabular}[c]{@{}c@{}}Prompt-Level \\ Accuracy\end{tabular} &
  0 &
  \{\{prompt\}\} \\
 &
  \rule{0pt}{21pt} FaithDial &
  Accuracy &
  8 &
  \begin{tabular}[c]{@{}c@{}}"Knowledge: " + \{\{knowledge\}\} + "\textbackslash{}nDialogue History: " + \\ \{\{history\_str\}\} + "\textbackslash{}nResponse: " + \{\{original\_response\}\} \\ + "\textbackslash{}nHallucinated:"\end{tabular} \\ 
 &
  \rule{0pt}{21pt} \begin{tabular}[c]{@{}c@{}}HaluEval \\ (QA)\end{tabular} &
  Accuracy &
  0 &
  \begin{tabular}[c]{@{}c@{}}"Knowledge: " + \{\{knowledge\}\} + "\textbackslash{}nQuestion: " \\ + \{\{question\}\} + "\textbackslash{}nAnswer: " + \{\{answer\}\} \\ + "\textbackslash{}nYour Judgement:"\end{tabular} \\ \bottomrule
\end{tabular}
}
\caption{Additional experimental setting details for the tasks. The double curly braces "\{\{\}\}" signify input data.}
\label{table:additional_details}
\end{table*}

\section{Experiment Settings}
\label{appendix:additional_details}
\autoref{table:additional_details} displays the additional experimental settings for the tasks considered in our leaderboard. Unless otherwise specified, we utilised the default settings provided by the EleutherAI Language Model Evaluation Harness \cite{eval-harness} framework.

% \section{Author Contributions}
% \label{appendix:author_contribution}

% GH conducted the first draft of the paper and the analyses in Section \ref{sec:comprehensive_alaysis}. APG contributed to the first version of the leaderboard and corresponding \href{https://huggingface.co/blog/leaderboard-hallucinations}{blog post}. RS contributed to the summarisation tasks. XD contributed to the knowledge memorisation tasks. LPB contributed to the writing and experimental design. MR contributed to prompt robustness evaluation and writing. PM created the first version of the leaderboard and corresponding \href{https://huggingface.co/blog/leaderboard-hallucinations}{blog post}.

\end{document}